\documentclass{article}
\usepackage{wrapfig}
\usepackage{lipsum}
\usepackage{tikz}
\usepackage{hyperref}
\usepackage{url}
\usepackage{amsfonts}
\usepackage{fancyhdr}
\usepackage{comment}
\usepackage{times}
\usepackage{float}
\usepackage{subfigure}
\usepackage{amsmath}
\usepackage{changepage}
\usepackage{amssymb}
\usepackage{graphicx}%
\usepackage{gensymb}
\usepackage{mhchem}
\usepackage{amsmath}

\usepackage{times}
\usepackage{graphicx} 
\usepackage{subfigure} 

\usepackage{natbib}

\usepackage{algorithm}
\usepackage{algorithmic}

\usepackage{hyperref}

\usepackage[accepted]{icml2016}

\icmltitlerunning{Online and Distributed learning of Gaussian mixture models by Bayesian Moment Matching}

\begin{document} 

\twocolumn[
\icmltitle{Online and Distributed learning of Gaussian Mixture Models \\ by Bayesian Moment Matching}

\icmlauthor{Priyank Jaini}{pjaini@uwaterloo.ca}
\icmladdress{David R. Cheriton School of Computer Science, University of Waterloo}
\icmlauthor{Pascal Poupart}{ppoupart@uwaterloo.ca}
\icmladdress{David R. Cheriton School of Computer Science, University of Waterloo}

\icmlkeywords{boring formatting information, machine learning, ICML}

\vskip 0.3in]

\begin{abstract} 
The Gaussian mixture model is a classic technique for clustering and data modeling that is used in numerous applications. With the rise of big data, there is a need for parameter estimation techniques that can handle streaming data and distribute the computation over several processors.  While online variants of the Expectation Maximization (EM) algorithm exist, their data efficiency is reduced by a stochastic approximation of the E-step and it is not clear how to distribute the computation over multiple processors.  We propose a Bayesian learning technique that lends itself naturally to online and distributed computation. Since the Bayesian posterior is not tractable, we project it onto a family of tractable distributions after each observation by matching a set of sufficient moments.  This Bayesian moment matching technique compares favorably to online EM in terms of time and accuracy on a set of data modeling benchmarks.
\end{abstract} 
\section{Introduction}

Gaussian Mixture models (GMMs)~\cite{murphy2012machine} are simple, yet expressive distributions that are often used for soft clustering and more generally data modeling.  Traditionally, the parameters of GMMs are estimated by batch Expectation Maximization (EM)~\cite{dempster1977maximum}. However, as datasets get larger and do not fit in memory or are continuously streaming, several online variants of EM have been proposed~\cite{Titterington,neal1998view,cappe2009line,liang2009online}. They process the data in one sweep by updating a sufficient statistics in constant time after each observation, however this update is approximate and stochastic, which slows down the learning rate.  Furthermore it is not clear how to distribute the computation over several processors given the sequential nature of those updates.

We propose a new Bayesian learning technique that lends itself naturally to online and distributed computation.  As pointed out by~\cite{broderick2013streaming}, Bayes' theorem can be applied after each observation to update the posterior in an online fashion and a dataset can be partitioned into subsets that are each processed by different processors to compute partial posteriors that can be combined into a single exact posterior that corresponds to the product of the partial posteriors divided by their respective priors.  

The main issue with Bayesian learning is that the posterior may not be tractable to compute and represent.  If we start with a prior that consists of the product of a Dirichlet by several Normal-Wisharts (one per Gaussian component) over the parameters of the GMM, the posterior becomes a mixture of products of Dirichlets by Normal-Wisharts where the number of mixture components grows exponentially with the number of observations. To keep the computation tractable, we project the posterior onto a single product of a Dirichlet with Normal-Wisharts by matching a set of moments of the approximate posterior with the moments of the exact posterior.  While moment matching is a popular frequentist technique that can be used to estimate the parameters of a model by matching the moments of the empirical distribution of a dataset~\cite{DBLP:journals/jmlr/AnandkumarHK12}, here we use moment matching in a Bayesian setting to project a complex posterior onto a simpler family of distributions. For instance, this type of Bayesian moment matching has been used in Expectation Propagation~\cite{minka2002expectation}.

Despite the approximation induced by the moment matching projection, the approach compares favorably to Online EM in terms of time and accuracy.  Online EM requires several passes through the data before converging and therefore when it is restricted to a single pass (streaming setting), it necessarily incurs a loss in accuracy while Bayesian moment matching converges in a single pass.  The approximation due to moment matching also induces a loss in accuracy, but the empirical results suggest that it is less important than the loss incurred by online EM.  Finally,  BMM lends itself naturally to distributed computation, which is not the case for Online EM.

The rest of the paper is structured as follows. Section~\ref{sec:Motivation} discusses the problem statement and motivation for online Bayesian Moment Matching algorithm. In Section~\ref{sec:background}, we give a brief background about the moment of methods and describe the family of distributions - Dirichlet, Normal-Wishart and Normal-Gamma, used as priors in this work. We further review the other online algorithm - online EM, used for parameter estimation of Gaussian Mixture models. Section~\ref{BMM} presents the Bayesian Moment matching algorithm for approximate Bayesian learning using moment matching. Section~\ref{sec:experiments} demonstrates the effectiveness of online BMM and online Distributed Moment Matching over online EM through empirical results on both synthetic and real data sets. Finally, Section~\ref{sec : conclusion} concludes the paper and talks about future work.

\section{Motivation}{\label{sec:Motivation}}
Given a set of data instances, where each data instance is assumed to be sampled independently and identically from a Gaussian mixture model, we want to estimate the parameters of the Gaussian mixture model in an online setting. 

More precisely, let $\textbf{x}^{1:N} = \{\textbf{x}_1, \textbf{x}_2,...,\textbf{x}_N\}$ be a set of \textit{n} data points, where each data point is sampled from a Gaussian mixture model with $M$ components. Let the parameters of this underlying Gaussian mixture model be denoted by $\Theta$, where $\Theta = \{\theta_1, \theta_2,....,\theta_{M}\}$. Each $\theta_i$ is a tuple of $(w_i,\mu_i, \Sigma_i) \; \forall i \in \{1,2,.....,M\}$ where $w_i$ is the weight, $\mu_i$ is the mean and $\Sigma_i$ is the covariance matrix of the $i^{th}$ component in the Gaussian mixture model. This can be expressed as 
\begin{equation*}
\textbf{x}_n \sim \sum_{i=1}^{M}w_i\mathcal{N}_{d}\big(\mu_i,\Sigma_i\big)
\end{equation*}
where \textit{d} denotes a d-dimensional Gaussian distribution and $\sum_{i=1}^Mw_i =1$ . The aim is to find an estimate $\hat{\Theta}$ of $\Theta$ in an online manner given the data $\textbf{x}^{1:N}$. \par 

One way to find the estimate $\hat{\Theta}$ is to compute the posterior $\textit{P}_n(\Theta) = Pr(\Theta | \textbf{x}^{1:n})$ by using Bayes theorem recursively. 
\begin{equation}\label{bayes exponential}
\begin{aligned}[b]
\textit{P}_n(\Theta) &= Pr(\Theta | \textbf{x}^{1:n}) \\
&\propto \textit{P}_{n-1}(\Theta)  Pr(\textbf{x}_n | \Theta) \\
&\propto Pr(\Theta | \textbf{x}^{1:n-1}) Pr(\textbf{x}_n | \Theta)\\
&= \frac{1}{k}Pr(\Theta | \textbf{x}^{1:n-1}) \sum_{i=1}^{M}w_i\mathcal{N}_{d}\big(\textbf{x}_n; \mu_i,\Sigma_i\big) 
\end{aligned}
\end{equation}
where k = $\int_{\Theta}Pr(\Theta | \textbf{x}^{1:n-1}) \sum_{i=1}^{M}w_i\mathcal{N}_{d}\big(\textbf{x}_n; \mu_i,\Sigma_i\big) d\Theta$ and the prior $\textit{P}_0 = f(\Theta| \Phi)$ be a distribution in $\Theta$ given parameter set $\Phi$. Hence, $\hat{\Theta} = \mathbb{E}[\textit{P}_N(\Theta)].$  \par

However, a major limitation with the approach above is that with each new data point $\textbf{x}_j$, the number of terms in the posterior given by Eq.~\ref{bayes exponential} increases by a factor $M$ due to the summation over the number of components. Hence, after \textit{N} data points, the posterior will consist of a mixture of $M^{N}$ terms, which is intractable. In this paper, we describe a Bayesian Moment Matching technique that helps to circumvent this problem. \par

The Bayesian Moment Matching (BMM) algorithm approximates the posterior obtained after each iteration in a manner that prevents the exponential growth of mixture terms in Eq.~\ref{bayes exponential}. This is achieved by approximating the distribution $\textit{P}_n(\Theta)$ obtained as the posterior by another distribution $\tilde{\textit{P}_n}(\Theta)$ which is in the same family of distributions $f(\Theta| \Phi)$ as the prior by matching a set of sufficient moments \textit{S} of $\textit{P}_n(\Theta)$ with $\tilde{\textit{P}_n}(\Theta)$. We will make this idea more concrete in the following sections.      
\section{Background}{\label{sec:background}}
\subsection{Moment Matching}{\label{Moment Matching}}
A moment is a quantitative measure of the shape of a distribution or a set of points. Let $\textit{f}(\boldsymbol\theta | \boldsymbol\phi)$ be a probability distribution over a d-dimensional random variable $\boldsymbol\theta = \{\theta_1, \theta_2,...,\theta_d \}$. The $j^{th}$ order moments of $\boldsymbol\theta$ are defined as $M_{g_j(\boldsymbol\theta)}(f) = \mathbb{E}\Big[ \prod_i\theta_i^{n_i}\Big]$ where $\sum_i n_i = j$ and $g_j$ is a monomial of $\boldsymbol\theta$ of degree \textit{j}.
\begin{equation*}
M_{g_j(\boldsymbol\theta)}(f) = \int_{\boldsymbol\theta} g_j(\boldsymbol\theta)f(\boldsymbol\theta | \boldsymbol\phi)d\boldsymbol\theta 
\end{equation*}
For some distributions \textit{f}, there exists a set of monomials \textit{S(f)} such that knowing $M_g(f) \; \forall g\in S(f)$ allows us to calculate the parameters of \textit{f}. For example, for a Gaussian distribution $\mathcal{N}(x ; \mu, \sigma^2)$, the set of sufficient moments \textit{S(f)} = \{$x, x^2$\}. This means knowing $M_x$ and $M_{x^2}$ allows us to estimate the parameters $\mu$ and $\sigma^2$ that characterize the distribution. We use this concept called the method of moments in our algorithm. \par 

Method of Moments is a popular frequentist technique used to estimate the parameters of a probability distribution based on the evaluation of the empirical moments of a dataset. It has been previously used to estimate the parameters of latent Dirichlet allocation, mixture models and hidden Markov models \cite{DBLP:journals/jmlr/AnandkumarHK12}. Method of Moments or moment matching technique can also be used for a Bayesian setting by computing a subset of the moments of the intractable posterior distribution given by Eq.~\ref{bayes exponential}. Subsequently, another tractable distribution from a family of distributions that matches the set of moments can be selected as an approximation for the intractable posterior distribution. For Gaussian mixture models, we use the Dirichlet as a prior over the weights of the mixture and a Normal-Wishart distribution as a prior over each Gaussian component. We next give details about the Dirichlet and Normal-Wishart distributions, including their set of sufficient moments.  
\subsection{Family of Prior Distributions}
In Bayesian Moment Matching, we project the posterior onto a tractable family of distribution by matching a set of sufficient moments. To ensure scalability, it is desirable to start with a family of distributions that is a conjugate prior pair for a multinomial distribution (for the set of weights) and Gaussian distribution with unknown mean and covariance matrix. The product of a Dirichlet distribution over the weights with a Normal-Wishart distribution over the mean and covariance matrix of each Gaussian component ensures that the posterior is a mixture of products of Dirichlet and Normal-Wishart distributions. Subsequently, we can approximate this mixture in the posterior with a single product of Dirichlet and Normal-Wishart distributions by using moment matching. We explain this in greater detail in Section~\ref{BMM}, but first we describe briefly the Normal-Wishart and Dirichlet distributions along with some sets of sufficient moments.     
\subsubsection{Dirichlet Distribution}\label{sec:dirichlet distribution}
The Dirichlet  distribution is a family of multivariate continuous probability distributions over the interval [0,1]. It is the conjugate prior probability distribution for the multinomial distribution and hence it is a natural choice of prior over the set of weights $\textbf{w} = \{w_1, w_2,...,w_{M}\}$ of a Gaussian mixture model. A set of sufficient moments for the Dirichlet distribution is $S = \{(w_i, w_i^2) : \forall i \in \{1,2,...,M\}\}$. Let $\boldsymbol\alpha = \{\alpha_1, \alpha_2,....,\alpha_{M}\}$ be the parameters of the Dirichlet distribution over \textbf{w}, then
\begin{equation}\label{Beta Moments}
\begin{aligned}[b]
\mathbb{E}[w_i] &= \frac{\alpha_i}{\sum_j \alpha_j} \quad \forall i \in \{1,2,...,M\}\\
\mathbb{E}[w_i^2] &= \frac{(\alpha_i)(\alpha_i +1)}{\Big(\sum_j \alpha_j\Big)\Big(1 +\sum_j \alpha_j\Big)}\quad \forall i \in \{1,2,...,M\}
\end{aligned}
\end{equation}
\subsubsection{Normal Wishart Prior}{\label{sec : normal-wishart distribution}}
The Normal-Wishart distribution is a multivariate distribution with four parameters. It is the conjugate prior of a multivariate Gaussian distribution with unknown mean and covariance matrix~\cite{DegrootNWi}. This makes a Normal-Wishart distribution a natural choice for the prior over the unknown mean and precision matrix for our case. \par 

Let $\boldsymbol\mu$ be a \textit{d}-dimensional vector and $\boldsymbol\Lambda$ be a symmetric positive definite $d \times d$ matrix of random variables respectively.  Then, a Normal-Wishart distribution over ($\boldsymbol\mu, \boldsymbol\Lambda$) given parameters ($\boldsymbol\mu_0, \kappa, \textbf{W}, \nu$) is such that
$\boldsymbol\mu \sim \mathcal{N}_d\big(\boldsymbol\mu ; \boldsymbol\mu_0 , (\kappa\boldsymbol\Lambda)^{-1}\big)$ where $\kappa > 0$ is real, $\boldsymbol\mu_0 \in \mathbb{R}^d$ and $\boldsymbol\Lambda$ has a Wishart distribution given as $\boldsymbol\Lambda \sim \mathcal{W}(\boldsymbol\Lambda ; \textbf{W}, \nu)$ where $\textbf{W} \in \mathbb{R}^{d \times d}$ is a positive definite matrix and $\nu > d-1$ is real. The marginal distribution of $\boldsymbol\mu$ is a multivariate t-distribution i.e $\boldsymbol\mu | \boldsymbol\Lambda \sim t_{\nu -d +1}\big(\boldsymbol\mu ; \boldsymbol\mu_0, \frac{\textbf{W}}{\kappa(\nu -d +1)}\big)$. The univariate equivalent for the Normal-Wishart distribution is the Normal-Gamma distribution.\par

In Section~\ref{Moment Matching}, we defined \textit{S}, a set of sufficient moments to characterize a distribution. In the case of the Normal-Wishart distribution, we would require at least four different moments to estimate the four parameters that characterize it. A set of sufficient moments in this case is $S = \{ \boldsymbol\mu, \boldsymbol\mu\boldsymbol\mu^{T}, \boldsymbol\Lambda, \Lambda_{ij}^2\}$ where $\Lambda_{ij}^2$ is the $(i,j)^{th}$ element of the matrix $\boldsymbol\Lambda$. The expressions for sufficient moments are given by 
\begin{equation}\label{Normal Wishart Moments}
\begin{aligned}[b]
\mathbb{E}[\boldsymbol\mu] &= \boldsymbol\mu_0 \\
\mathbb{E}[(\boldsymbol\mu - \boldsymbol\mu_0)(\boldsymbol\mu - \boldsymbol\mu_0)^T] &= \frac{\kappa + 1}{\kappa(\nu - d -1)}\textbf{W}^{-1} \\
\mathbb{E}[\boldsymbol\Lambda] &= \nu\textbf{W} \\
Var(\Lambda_{ij}) &= \nu(W_{ij}^2 + W_{ii}W_{jj})
\end{aligned}
\end{equation}
\subsection{Online Expectation Maximization}
Batch Expectation Maximization~\cite{dempster1977maximum} is often used in practice to learn the parameters of the underlying distribution from which the given data is assumed to be derived. In~\cite{Titterington}, a first online variant of EM was proposed, which was later modified and improved in several variants~\cite{neal1998view,sato2000line,cappe2009line,liang2009online} that are closer to the original batch EM algorithm. In online EM, an updated parameter estimate $\hat{\boldsymbol\Theta_n}$ is produced after observing each data instance $\textbf{x}_n$. This is done by replacing the expectation step by a stochastic approximation, while the maximization step is left unchanged. In the limit, online EM converges to the same estimate as batch EM when it is allowed to do several iterations over the data.  Hence, a loss in accuracy is incurred when it is restricted to a single pass over the data as required in the streaming setting. 
\begin{algorithm}[!h]
   \caption{Generic Bayesian Moment Matching}
   \label{alg: general BMM}
\begin{algorithmic}
   \STATE {\bfseries Input:} Data $\textbf{x}_i$, $i \in \{1,2,...,N\}$
   \STATE Let $\textit{f}(\boldsymbol\Theta|\boldsymbol\Phi)$ be a family of probability distributions with 	parameters $\boldsymbol\Phi$ 
   \STATE Initialize a prior $\textit{P}_0(\boldsymbol\Theta)$
   \FOR{$n=1$ {\bfseries to} $N$} 
   \STATE Compute $\textit{P}_n(\boldsymbol\Theta)$ from $\textit{P}_{n-1}(\boldsymbol\Theta)$ using 	Eq.~\ref{bayes exponential}
   \STATE $\forall \; g(\boldsymbol\Theta) \in S(f)$, evaluate $M_{g(\boldsymbol\Theta)}(\textit{P}_n)$
   \STATE Compute $\boldsymbol\Phi$ using $M_{g(\boldsymbol\Theta)}(\textit{P}_n)$'s
   \STATE Approximate $\textit{P}_n$ with $\tilde{\textit{P}_n}(\boldsymbol\Theta) = \textit{f}(\boldsymbol\Theta|\boldsymbol\Phi)$
	 \ENDFOR
   \STATE Return $\hat{\boldsymbol\Theta} = \mathbb{E}[\tilde{\textit{P}_n}(\boldsymbol\Theta)]$
\end{algorithmic}
\end{algorithm}
\section{Bayesian Moment Matching}{\label{BMM}} 
We now discuss in detail the Bayesian Moment Matching (BMM) algorithm. BMM approximates the posterior after each observation with fewer terms in order to prevent the number of terms to grow exponentially. \par 
In Algorithm~\ref{alg: general BMM}, we first describe a generic procedure to approximate the posterior $\textit{P}_n$ after each observation with a simpler distribution $\tilde{\textit{P}_n}$ by moment matching. More precisely, a set of moments sufficient to define $\tilde{\textit{P}_n}$ are matched with the moments of the exact posterior $\textit{P}_n$. For every iteration, we first calculate the exact posterior $\textit{P}_n(\boldsymbol\Theta | \textbf{x}^{1:n})$. Then, we compute the set of moments \textit{S(f)} that are sufficient to define a distribution in the family $f(\boldsymbol\Theta | \boldsymbol\Phi)$. Next, we compute the parameter vector $\boldsymbol\Phi$ based on the set of sufficient moments. This determines a specific distribution $\tilde{\textit{P}_n}$ in the family $f$ that we use to approximate $\textit{P}_n$. Note that the moments in the sufficient set \textit{S(f)} of the approximate posterior are the same as that of the exact posterior. However, all the other moments outside this set of sufficient moments \textit{S(f)} may not necessarily be the same. \par

In the next section (\ref{sec : BMM - uGMM}), we illustrate Algorithm~\ref{alg: general BMM} for learning the parameters of a univariate Gaussian mixture model. Subsequently, we will give the BMM algorithm for general multivariate Gaussian mixture models. 

\subsection{BMM for univariate Gaussian mixture model}{\label{sec : BMM - uGMM}}
In this section, we illustrate the Bayesian moment matching algorithm for Gaussian mixture models. Let $x^{1:n}$ be a dataset of \textit{n} data points derived from a univariate Gaussian mixture model with density function given by $Pr(x|\Theta) = \sum_{i=1}^{M}w_i \mathcal{N}\big(x ; \mu_i, \sigma_i^2\big)$, where $\Theta = \{(w_1,\mu_1,\sigma_1^2),(w_2,\mu_2,\sigma_2^2),...(w_M,\mu_M,\sigma_M^2)\}$. \par
The first step is to choose an appropriate family of distributions $f(\Theta|\Phi)$ for the prior $\textit{P}_0(\Theta)$. A conjugate prior probability distribution pair of the likelihood $Pr(x|\Theta)$ would be a desirable family of distributions. We further make the assumption that every component of GMM are independent of all the other components. The independence assumption helps to simplify the expressions for the posterior. Hence, the prior is chosen as a product of a Dirichlet distribution over the weights $w_i$ and Normal-Gamma distributions over each tuple $(\mu_i, \lambda_i)$ where $\lambda_i = (\sigma_i^2)^{-1}$. More precisely, $\textit{P}_0(\Theta) = Dir(\textbf{w}|\textbf{a})\prod_{i=1}^{M}\mathcal{NG}(\mu_i, \lambda_i|\alpha_i,\kappa_i, \beta_i, \gamma_i)$ where $\textbf{w} = (w_1,w_2,...,w_M)$ and $\textbf{a} = (a_1,a_2,...,a_M)$.\par 
Given a prior $\textit{P}_0(\Theta)$, the posterior $\textit{P}_1(\Theta|x_1)$ after observing the first data point $x_1$ is given by
\begin{center}
$\textit{P}_1(\Theta|x_1) \propto \textit{P}_{0}(\Theta)  Pr(\textbf{x}_1 | \Theta)$
\end{center}
\begin{equation}\label{eq: example uGMM}
\begin{aligned}[b]
&=\frac{1}{k}Dir(\textbf{w}|\textbf{a})\prod_{i=1}^{M}\mathcal{NG}(\mu_i, \lambda_i|\alpha_i,\kappa_i, \beta_i, \gamma_i)\\ & \qquad \qquad \qquad  \qquad  \qquad \sum_{j=1}^{M}w_j\mathcal{N}\big(x_1; \mu_j,\sigma_j^2\big)\\ 
&= \frac{1}{k} \sum_{j=1}^{M}w_jDir(\textbf{w}|\textbf{a})\prod_{i=1}^{M}\mathcal{NG}(\mu_i, \lambda_i|\alpha_i,\kappa_i, \beta_i, \gamma_i)\\& \qquad \qquad \qquad \qquad \qquad  \qquad\mathcal{N}\big(x_1; \mu_j,\sigma_j^2\big)
\end{aligned}
\end{equation}

Since, a Normal-Gamma distribution is a conjugate prior for a Normal distribution with unknown mean and variance, $\mathcal{NG}(\mu_i, \lambda_i|\alpha_i,\kappa_i, \beta_i, \gamma_i)\mathcal{N}\big(x_1; \mu_i,\sigma_i^2\big) = c\mathcal{NG}(\mu_i, \lambda_i|\alpha_i^*,\kappa_i^*, \beta_i^*, \gamma_i^*)$ where $c$ is some constant. Similarly, $w_iDir(w_1,w_2,...,w_M|a_1,a_2,..,a_i,..,a_M) = uDir(w_1,w_2,...,w_M|a_1,a_2,..a_i^*..,a_M)$ where $u$ is some constant and
\begin{equation}\label{eq: updates uGMM}
\begin{aligned}[b]
& \alpha_i^{*} = \frac{\kappa_i \alpha_i + x_1}{\kappa_i + 1}\\
& \kappa_i^{*} = 1 + \kappa_i \\
& \beta_i^{*} = \beta_i + \frac{1}{2} \\
& \gamma_i^{*} = \gamma_i + \kappa_i\frac{(x_1 - \alpha_i)^{2}}{2(1 + \kappa_i)}\\
& c = \sqrt[2]{(\frac{\kappa_i}{\kappa_i^{*}})}\;\frac{\Gamma(\beta_i^{*})}{\Gamma(\beta_i)}\frac{(\gamma_i)^{(\beta_i)}}{(\gamma_i^{*})^{(\beta_i^{*})}}\\
& a_i^* = a_i +1
\end{aligned}
\end{equation}

Therefore, Eq.~\ref{eq: example uGMM} can now be expressed as
\begin{multline}{\label{eq:updated posterior uBMM}}
\textit{P}_1(\Theta|x_1) = \sum_{j=1}^{M}\Bigg(c_jDir(\textbf{w}|\textbf{a}_j^{*})\mathcal{NG}(\mu_j, \lambda_j|\alpha_j^*,\kappa_j^*, \beta_j^*, \gamma_j^*)\\ \prod_{i\neq j}^{M}\mathcal{NG}(\mu_i, \lambda_i|\alpha_i,\kappa_i, \beta_i, \gamma_i)\Bigg)
\end{multline}
where $\textbf{a}_j^* = (a_1,a_2,..,a_j^*,..,a_M)$ and k is the normalization constant. Eq.~\ref{eq:updated posterior uBMM} suggests that the posterior is a mixture of product of distributions where each product component in the summation has the same form as that of the family of distributions of the prior $\textit{P}_0(\Theta)$. It is evident from Eq.~\ref{eq:updated posterior uBMM} that the terms in the posterior grow by a factor of \textit{M} for each iteration, which is problematic. In the next step, we approximate this mixture $P_1(\Theta)$ with a single product of Dirichlet and Normal-Gamma distributions $\tilde{P_1}(\Theta)$ by matching all the sufficient moments of $P_1$ with $\tilde{P_1}$ i.e.
\begin{equation}\label{eq: mm uBMM}
\begin{aligned}[b]
&\tilde{P}_1(\Theta) \simeq P_1(\Theta) \\
where &\quad \tilde{P}_1(\Theta) = Dir(\textbf{w}|\textbf{a}^1)\prod_{i=1}^{M}\mathcal{NG}(\mu_i, \lambda_i|\alpha_i^1,\kappa_i^1, \beta_i^1, \gamma_i^1)
\end{aligned}
\end{equation}
We evaluate the parameters $\textbf{a}^1,\alpha_i^1,\kappa_i^1, \beta_i^1, \gamma_i^1$ by matching some sufficient moments of $\tilde{P}_1(\Theta)$ with  $P_1(\Theta)$. The set of sufficient moments for the posterior is $S(P_1) = \{(\mu_j,\lambda_j,\lambda_j^2,\mu_j\lambda_j^2,w_j,w_j^2)\; |\; \forall j \in 1,2,...,M \}$. For any $g \in S(P_1)$
\begin{equation}
\mathbb{E}[g] = \int_{\Theta} g P_1(\Theta) d(\Theta)
\end{equation}
The parameters of $\tilde{P_1}$ can be computed from the following set of equations
\begin{equation}\label{eq: posterior parameters}
\begin{aligned}[b]
\alpha_j^1 &= \mathbb{E}[\mu_j] \\
\kappa_j^1 &= \frac{1}{\mathbb{E}[\mu_j\lambda_j^2] - \mathbb{E}[\mu_j]^2\mathbb{E}[\lambda_j]}\\
\beta_j^1 &= \frac{\mathbb{E}[\lambda_j]^2}{\mathbb{E}[\lambda_j^2] - \mathbb{E}[\lambda_j]^{2}} \\
\gamma_j^1 &= \frac{\mathbb{E}[\lambda_j]}{\mathbb{E}[\lambda_j^2]- \mathbb{E}[\lambda_j]^{2}}  \\
a_j^1 &= \mathbb{E}[w_j]\frac{\mathbb{E}[w_j] - \mathbb{E}[w_j^2]}{\mathbb{E}[w_j^2] - \mathbb{E}[w_j]^2}\quad \forall j \in \{1,2,...,M\}
\end{aligned}
\end{equation}
Using the set of equations given by (\ref{eq: posterior parameters}), we approximate the exact posterior $P_1(\Theta)$ with $\tilde{P_1}(\Theta)$. This posterior will be the prior for the next iteration and we keep following the steps above iteratively to finally have a distribution $\tilde{P_n}(\Theta)$ after observing a stream of data $x^{1:n}$. The estimate $\hat{\Theta} = \mathbb{E}[\tilde{P_n}(\Theta)]$ is returned. \par
Here, we have assumed that the number of components $M$ is known. In practice, however, this may not be the case. This problem can be addressed by taking a large enough value of $M$ while learning the model. Although, such an approach might lead to overfitting for maximum likelihood techniques such as online EM, in our case, this is a reasonable approach since Bayesian learning is fairly robust to overfitting. 
\subsection{BMM for multivariate Gaussian mixture model}
In the previous section, we illustrated the Bayesian moment matching algorithm for a univariate Gaussian mixture model in detail. In this section we briefly discuss the general case for a multivariate Gaussian mixture model. The family of distributions for the prior $P_0(\boldsymbol\Theta)$ in this case becomes $P_0(\boldsymbol\Theta) = Dir(\textbf{w}|\textbf{a})\prod_{i=1}^{M}\mathcal{NW}_d(\boldsymbol\mu_i, \boldsymbol\Lambda_i|\boldsymbol\alpha_i,\kappa_i, \textbf{W}_i, \nu_i)$ where $\textbf{w} = (w_1,w_2,...,w_M)$ and $\textbf{a} = (a_1,a_2,...,a_M)$. The algorithm works in the same manner as shown before. However, the update equations in (\ref{eq: updates uGMM}) would now change accordingly. \par
The set of sufficient moments for the posterior in this case would be given by $S(P(\boldsymbol\Theta|\textbf{x}))=\{\boldsymbol\mu_j, \boldsymbol\mu_j\boldsymbol\mu_j^T, \boldsymbol\Lambda_j,\Lambda_{j_{kl}}^2, w_j, w_j^2 : \forall j \in 1,2,..,M\}$ where $\Lambda_{j_{kl}}$ is the $(k,l)^{th}$ element of the matrix $\boldsymbol\Lambda_j$. Notice that, since $\boldsymbol\Lambda_j$ is a symmetric matrix, we only need to consider the moments of the elements on and above the diagonal of $\boldsymbol\Lambda_j$. \par

In Eq.~\ref{Normal Wishart Moments} of Section~\ref{sec : normal-wishart distribution}, we presented the expressions for a set of sufficient moments of a Normal-Wishart distribution. Using those expressions we can again approximate a mixture of products of Dirichlet and Normal-Wishart distributions in the posterior with a single product of Dirichlet and Normal-Wishart distributions, as we did in the previous section. Finally, the estimate $\boldsymbol\Theta = \mathbb{E}[\tilde{P_n}(\boldsymbol\Theta)]$ is obtained after observing the data $\textbf{x}^{1:n}$. In Algorithm~\ref{alg:BMM for GMM}, we give the algorithm for Bayesian moment matching for Gaussian mixture models.
\begin{algorithm}[tb]
   \caption{Bayesian Moment Matching for Gaussian mixture}
   \label{alg:BMM for GMM}
\begin{algorithmic}
   \STATE {\bfseries Input:} Data $\textbf{x}_i$, $i \in \{1,2,...,N\}$
   \STATE Let $\textit{f}(\boldsymbol\Theta|\boldsymbol\Phi)$ be a family of probability distributions given by a product of a Dirichlet and Normal-Wishart distributions. 
   \STATE Initialize $\textit{P}_0(\boldsymbol\Theta)$ as $Dir(\textbf{w}|\textbf{a})\prod_{i=1}^{M}\mathcal{NW}_d(\boldsymbol\mu_i, \boldsymbol\Lambda_i|\boldsymbol\alpha_i,\kappa_i, \linebreak \textbf{W}_i, \nu_i)$
   \FOR{$n=1$ {\bfseries to} $N$} 
   \STATE Compute $\textit{P}_n(\boldsymbol\Theta)$ from $\textit{P}_{n-1}(\boldsymbol\Theta)$ using 	Eq.(\ref{bayes exponential})
   \STATE $\forall \; g(\boldsymbol\Theta) \in S(f)$, evaluate $M_{g(\boldsymbol\Theta)}(\textit{P}_n)$
   \STATE Compute $\boldsymbol\Phi$ using $M_{g(\boldsymbol\Theta)}(\textit{P}_n)$'s
   \STATE Approximate $\textit{P}_n$ with $\tilde{\textit{P}_n}(\boldsymbol\Theta) = \textit{f}(\boldsymbol\Theta|\boldsymbol\Phi)$
	 \ENDFOR
   \STATE Return $\hat{\boldsymbol\Theta} = \mathbb{E}[\tilde{\textit{P}_n}(\boldsymbol\Theta)]$
\end{algorithmic}
\end{algorithm}
\subsection{Distributed Bayesian Moment Matching}
One of the major advantages of Bayes' theorem is that the computation of the posterior can be distributed over several machines, each of which processes a subset of the data. It is also possible to compute the posterior in a distributed manner using Bayesian moment matching algorithm. For example, let us assume that we have \textit{T} machines and a data set with \textit{TN} data points. Each machine $t$, can compute the approximate posterior $P_t(\boldsymbol\Theta|\textbf{x}^{(t-1)N+1:tN})$ where $t\in 1,2,..,T$ using Algorithm~\ref{alg:BMM for GMM} over \textit{N} data points. These partial posteriors $\{P_t\}_{t=1}^T$ can be combined to obtain a posterior over the entire data set $\textbf{x}^{1:TN}$ according to the following equation:
\begin{align}
P(\boldsymbol\Theta|\textbf{x}^{1:TN}) = P(\boldsymbol\Theta) \prod_{t=1}^{T}\frac{P_t(\boldsymbol\Theta|\textbf{x}^{(t-1)N+1:tN})}{P(\boldsymbol\Theta)}
\end{align}
Subsequently, the estimate $\hat{\boldsymbol\Theta}=\mathbb{E}[P(\boldsymbol\Theta|\textbf{x}^{1:TN})]$ is obtained over the whole data set. Therefore, we can use Bayesian moment matching algorithm to perform Bayesian learning in an online and distributed fashion. We will show in Section~\ref{sec:experiments} that distributed Bayesian moment matching performs favorably in terms of accuracy and results in a huge speed-up of running time.
\section{Experiments}{\label{sec:experiments}}
We performed experiments on both synthetic and real datasets to evaluate the performance of online Bayesian moment matching algorithm (oBMM). We used the synthetic datasets to verify whether oBMM converged to the true model given enough data. We subsequently compared the performance of oBMM with the online Expectation Maximization algorithm (oEM) described in~\cite{cappe2009line}. We compared oBMM with this version of oEMsince it has been shown to perform best among various variants of oEM~\cite{liang2009online}. We now discuss experiments on both kinds of datasets in detail.
\subsection*{Synthetic Data sets}
\begin{figure*}[ht]
\vskip 0.2in
\begin{center}
\centerline{\includegraphics[width= 15cm,height=13cm,keepaspectratio]{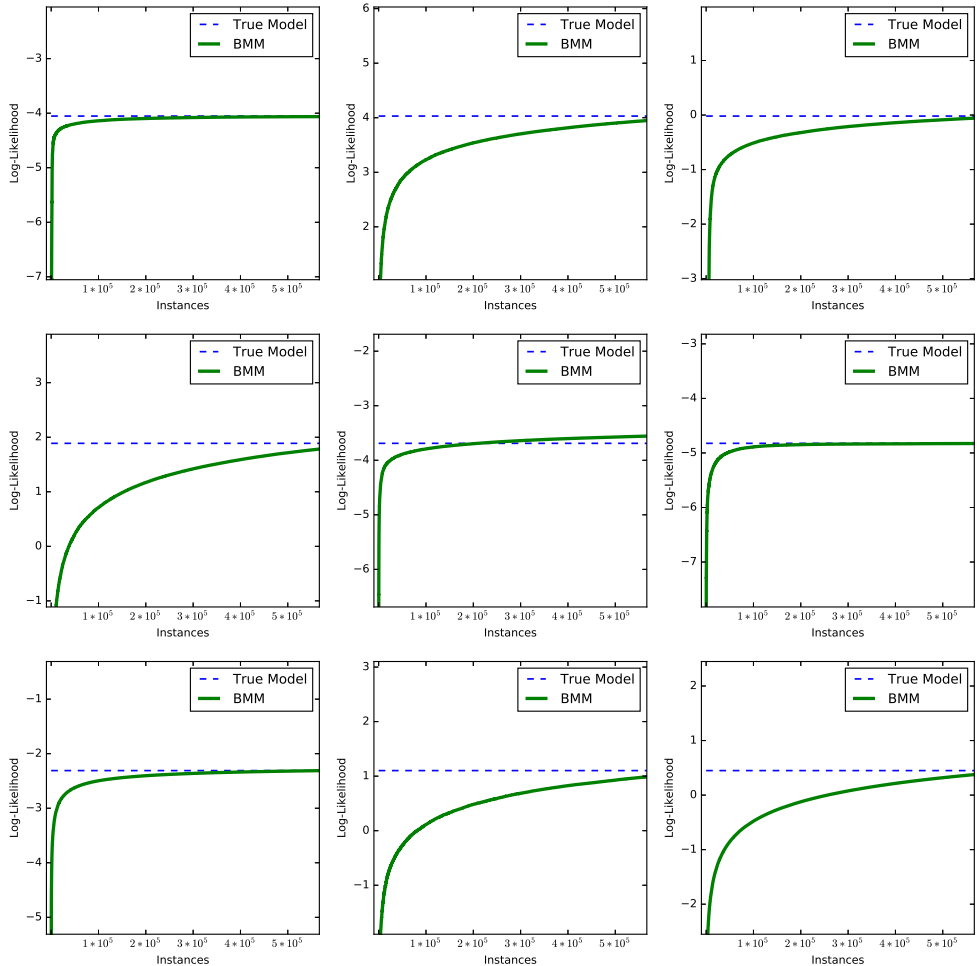}}
\caption{Performance Analysis of online Bayesian moment matching algorithm for Gaussian mixture models on synthetic datasets with 170,000 training instances and 30,000 testing instances. The plot shows the convergence of log-likelihood of the model learned by BMM v\/s number of observed data instances. The plot clearly shows convergence to the true model.}
\label{fig : log-likelihood plots}
\end{center}
\vskip -0.2in
\end{figure*}

We evaluate the performance of oBMM on 9 different synthetic data sets. All the data sets were generated with a Gaussian mixture model with a different number of components lying in the range of 2 to 6 components and having a different number of attributes (or dimensions) in the range of 3 to 10 dimensions. For each data set, we sampled 200,000 data points. We divided each data set in to a training set with 170,000 data instances and 30,000 testing instances. To evaluate the performance of oBMM, we calculated the average log-likelihood of the model learned by oBMM after each data instance is observed. Figure~\ref{fig : log-likelihood plots} shows the plots for performance of oBMM against the true model. Each subplot has the average log-likelihood on the vertical axis and the number of observation on the horizontal axis. It is clear from the plots, that oBMM converges to the true model likelihood, in each of the nine cases, given a large enough data set. 
\par   
Next, we discuss the performance of oBMM against oEM and show through experiments on real data sets that oBMM performs better than oEM in terms of both accuracy and running time. 
\subsection*{Real Data sets}
We evaluated the performance of oBMM on 2 sets of real datasets - 10 moderate-small size datasets and 4 large datasets available publicly online at the UCI machine learning repository and Function Approximation repository\cite{Datset}. All the datasets span over diverse domains. The number of attributes(or dimensions) range from 4 to 91. \par
In order to evaluate the performance of oBMM, we compare it to oEM. We measure both - the quality of the two algorithms in terms of average log-likelihood scores on the held-out test datasets and their scalability in terms of running time. We use the Wilcoxon signed ranked test\cite{wilcoxon1950some} to compute the \textit{p}-value and report statistical significance with \textit{p}-value less than 0.05, to test the statistical significance of the results. We computed the parameters for each algorithm over a range of components varying from 2 to 10. For analysis, we report the model for which the log-likelihood over the test data stabilized and showed no further significant improvement for both oEM and oBMM. For oEM the step size for the stochastic approximation in the E-Step was set to $(n+3)^{-\alpha}$ where $0.5\leq\alpha\leq1$\cite{liang2009online} where \textit{n} is the number of observations. We evaluate the performance of online Distributed Moment Matching (oDMM) by dividing the training datasets in to 5 smaller data sets, and processing each of these small datasets on a different machine. The output from each machine is collected and combined to give a single estimate for the parameters of the model learned.  \par
\begin{table}[!h]
\caption{Log-likelihood scores on 10 data sets. The best results among oBMM and oEM are highlighted in bold font. $\uparrow$(or $\downarrow$) indicates that the method has significantly better (or worse) log-likelihoods than Online Bayesian Moment Matching (oBMM) under Wilcoxon signed rank test with pvalue $<$ 0.05.}
\label{table : log-likelihood small datasets}
\vskip 0.15in
\begin{center}
\begin{small}
\begin{sc}
\begin{tabular}{l@{\hspace{3mm}}r@{\hspace{3mm}}r@{\hspace{3mm}}r@{\hspace{3mm}}r}
\hline
\abovespace\belowspace
Data set & Instances & oEM & oBMM \\
\hline
\abovespace
Abalone    & 4177& -2.65 $\downarrow$& \textbf{-1.82}$$ \\
Banknote   & 1372 & -9.74 $\downarrow$ &\textbf{-9.65} \\
Airfoil    & 1503 & \textbf{-15.86}\;\;\; & -16.53 \\
Arabic     & 8800 & -15.83 $\downarrow$ & \textbf{-14.99}\\
Transfusion& 748 & -13.26 $\downarrow$ & \textbf{-13.09}    \\
CCPP       & 9568 & -16.53 $\downarrow$ & \textbf{-16.51}\\
Comp. Activity &8192 & -132.04 $\downarrow$ & \textbf{-118.82}     \\
Kinematics     & 8192 & -10.37 $\downarrow$ & \textbf{-10.32}\\
Northridge     &2929 & -18.31 $\downarrow$ & \textbf{-17.97}\\
\belowspace
Plastic      & 1650 & -9.4694 $\downarrow$ & \textbf{-9.01}\\
\hline
\end{tabular}
\end{sc}
\end{small}
\end{center}
\vskip -0.1in
\end{table}
 
Table~\ref{table : log-likelihood small datasets} shows the average log-likelihood on test sets for oBMM and oEM. oBMM outperforms oEM on 9 of the 10 datasets. The results show that for some datasets, oBMM has significantly better log-likelihoods than oEM. Table~\ref{table : log-likelihood big datasets} and Table~\ref{table : Running time} show the log-likelihood scores and running times of each algorithm on large datasets. In terms of log-likelihood scores, oBMM outperforms oEM and oDMM on all 4 datasets. While, the performance of oDMM is expected to be worse than oBMM, it is to be noticed that the performance of oDMM is not very significantly worse. This is encouraging in light of the huge gains in terms of running time of oDMM over oEM and oBMM. Table~\ref{table : Running time} shows the performance of each algorithm in terms of running times. oDMM outperforms each of the other algorithms very significantly. It is also worth noting that oBMM performed better than oEM on 3 out of 4 datasets.

\begin{table}[!h]
\caption{Log-likelihood scores on 4 large data sets. The best results among oBMM, oDMM and oEM are highlighted in bold font.}
\label{table : log-likelihood big datasets}
\vskip 0.15in
\begin{center}
\begin{small}
\begin{sc}
\begin{tabular}{@{}l@{\hspace{2mm}}r@{\hspace{2mm}}r@{\hspace{2mm}}r@{\hspace{2mm}}r@{}}
\hline
\abovespace\belowspace
Data (Attributes) & Instances & oEM & oBMM & oDMM \\
\hline
\abovespace
Heterogeneity (16) & 3930257 & -176.2 & \textbf{-174.3} & -180.7\\
Magic 04 (10) & 19000 & -33.4 & \textbf{-32.1} & -35.4 \\
Year MSD (91)& 515345 & -513.7 & \textbf{-506.5}& -513.8 \\
\belowspace
MiniBooNe (50) & 130064 &-58.1&\textbf{-54.7}&-60.3
\end{tabular}
\end{sc}
\end{small}
\end{center}
\vskip -0.1in
\end{table}

\begin{table}[!h]
\caption{Running time in seconds on 4 large datasets. The best running time is highlighted in bold fonts}
\label{table : Running time}
\vskip 0.15in
\begin{center}
\begin{small}
\begin{sc}
\begin{tabular}{@{}l@{\hspace{2mm}}r@{\hspace{2mm}}r@{\hspace{2mm}}r@{\hspace{2mm}}r@{}}
\hline
\abovespace\belowspace
Data (Attributes) & Instances & oEM & oBMM & oDMM \\
\hline
\abovespace
Heterogeneity (16) & 3930257 & 77.3 & 81.7 & \textbf{17.5}\\
Magic 04 (10) & 19000 & 7.3 & 6.8 & \textbf{1.4} \\
Year MSD (91)& 515345 & 336.5 & 108.2& \textbf{21.2} \\
\belowspace
MiniBooNe (50) & 130064 &48.6 & 12.1 & \textbf{2.3}
\end{tabular}
\end{sc}
\end{small}
\end{center}
\vskip -0.1in
\end{table}
\section{Conclusion}{\label{sec : conclusion}}
With the advent of technology, large data sets are being generated in almost all fields - scientific, social, commercial - spanning diverse areas like physics, molecular biology, social networks, health care, trading markets, to name a few. Therefore, it has become imperative to develop algorithms which can process these large data sets in minimum time in an online fashion. In this paper, we explored online algorithms to learn the parameters of Gaussian Mixture models. We proposed an online Bayesian Moment Matching algorithm for parameter learning and demonstrated how it can be used in a distributed manner leading to substantial gains in running time. We further showed through empirical analysis that the online Bayesian Moment Matching algorithm converges to the true model and outperforms online EM both in terms of accuracy and running time. We also demonstrated that distributing the algorithm over several machines results in faster running times without significantly compromising accuracy, which is particularly advantageous when running time is a major bottleneck. \par
In the future, we would like to further develop the online Bayesian Moment Matching algorithm to learn the number of components in a mixture model in an online fashion. Some work has already been done in this direction with Dirichlet process mixtures~\cite{wang2012truncation,lin2013online} and it would be desirable to explore how the BMM algorithm can be adapted to learn the number of components. Further, we can use the proposed online BMM for Gaussian Mixture models to extend this work to learn a Sum-Product Network with continuous variables in an online manner.
\bibliography{reference}
\bibliographystyle{icml2016}
\end{document}